\documentclass[12pt]{article}
\usepackage{graphicx,times,amsmath,amssymb} 
\newcommand{\ee}{\end{equation}}             
\newcommand{\ba}{\begin{array}}              
\newcommand{\ea}{\end{array}}                
\newcommand{\bea}{\begin{eqnarray}}          
\newcommand{\eea}{\end{eqnarray}}            
\newcommand{\bas}{\begin{eqnarray*}}
\newcommand{\eas}{\end{eqnarray*}}
\newcommand{\beqa}{\be \begin{array}{rcl}}   
\newcommand{\eeqa}{\end{array} \ee}          
\newcommand\Fin{\hfill $\diamond $}          
\newtheorem{Def}{\bf Definition}
\newtheorem{Obs}[Def]{\bf Remark}
\newtheorem{Ex}[Def]{\bf Example}
\newtheorem{Prop}[Def]{\bf Proposition}
\newtheorem{Lemm}[Def]{\bf Lemma}
\newtheorem{Cor}[Def]{\bf Corollary}
\newtheorem{Obss}[Def]{\bf Remarks}
\newcommand{\sect}[1]{\setcounter{equation}{0} \section{#1}} 
\newcommand{\bld}[1]{\mbox{\boldmath $#1$}}

\newcommand{\half}{{\textstyle \frac{1}{2}}}            
\newcommand{\bhalf}{ \scriptstyle \frac{1}{2}}                
\newcommand{\sqr}{{\textstyle \sqrt{2}}}                

\newcommand{\twoi}{\scriptstyle \sqrt{2}}               
\newcommand{\sqtwo}{{\textstyle \frac{1}{\sqrt{2}}}}    

\newcommand{\sci}{{\scriptstyle \circ}}                 
\newcommand{\und}{\underline}                           

\newcommand\Rr{\mathbb R}            
\newcommand\Cl{\it Cl}               
\newcommand\Ge{\cal G}               
\newcommand\Co{\cal V}               
\newcommand\Ok{\cal O}               
\newcommand\Af{\mathbb A}            
\newcommand\Pn{\mathbb P}            
\newcommand\Ps{\mathbb P \mathbb R}  
\newcommand\No{\mathbb N \mathbb O}  
\newcommand\Ni{\mathbb N \mathbb I}  
\newcommand\In{\cal I}               
\newcommand\Pk{\mathbb P \mathbb K}  
\begin{document}
\title{\Large \bf Clifford Algebra of the Vector Space of Conics for decision boundary Hyperplanes in $m$-Euclidean Space. }
\author{Isidro Nieto  \thanks{The  first author is  very thankful for being allowed to use the facilities of the Economy School in writing this paper.} ~~~~J. Refugio Vallejo \thanks{Supported by Concyteg Project GTO-04-C02-93.}      \\ 
e-mail:{\tt inieto@quijote.ugto.mx~,~cuco@quijote.ugto.mx } \\ 
Geomatic and Hydraulic Engineering,~University of Guanajuato, \\ Av.~Juarez No.~77, C.P.~36000, Guanajuato,~Gto.,~M\'exico \\
and \\ 
 Economy School,~University of Guanajuato, \\ 
 UCEA-Campus Marfil,~Fracc. I,~El  Establo,~C.P.~36250,~Guanajuato, Gto.,~M\'exico}

\maketitle


\begin{abstract}
In this paper we embed $m$-dimensional Euclidean space in the geometric algebra $\Cl_m $ to extend the operators of incidence in ${\Rr^m}$ to operators of incidence in the geometric algebra to generalize the notion of separator to a decision boundary hyperconic in the Clifford algebra of hyperconic sections  denoted as ${\Cl}( {\Co}_{2})$. This allows us to extend the concept of a linear perceptron or the spherical perceptron in conformal geometry  and introduce  the more general conic perceptron, namely the {\it elliptical perceptron}.~Using  Clifford duality  a vector orthogonal to the decision boundary hyperplane is determined.~Experimental results  are shown in 2-dimensional Euclidean space where we separate data that are  naturally separated by some typical plane conic separators by this procedure. This procedure is more general in the sense that it is  independent of the dimension of the input data and hence we can speak of the  hyperconic elliptic perceptron.
\end{abstract}


\vspace{0.5cm}
\section{Introduction}
In this paper we extend the operators of incidence in $m$-dimensional Euclidean space to operators of incidence in the geometric algebra to take the advantage of simple representation of geometric entities on the one hand and its low computational complexity on the other.
More concretly, in the case of linear subspaces of $m$-dimensional Euclidean space , the perceptrons such as the hyperplanes, hyperspheres and hyperconic find a representation as hyperplanes or linear subspaces in the geometric algebra where the notions of incidence are exactly expressed as in the case of $m$-dimensional Euclidean space. In the simplest case, which is the linear perceptron the input data is divided in two classes of points by the hyperplane, namely the points of one side and the points of the other side of it. Another example is that of the spherical perceptron where we define two classes of input data which are  the interior points and exterior points to the circle. In the latter, the circle is in fact represented as a hyperplane in the conformal geometric algebra ${\Pk}^m$.  
 
 In the same way that the lines, planes or hyperplanes are the simplest and natural separators in the Euclidean space ${\Rr}^m$ of two classes where we define the perceptron, the circle, the sphere and hypersphere are the natural separators of two classes in the conformal space ${\Pk}^m$. In this space we define spherical perceptron and also the spherical neural networks in which we can separate points from structures that have an interior and exterior.

Similarly as  the perceptron  is defined to separate linearly two classes and the spherical perceptron to separate spherically interior from exterior we define the elliptic perceptron to separate points of one side of the conic and points in the other side which have a conic as a natural boundary decision hypersurface; it includes hyperplanes, hyperspheres, hyperellipses and hyperbolic surfaces; with this separator we generalize any other separators. We also use this conic separator to extend the concept of spherical neural network to define the elliptic neural network which is a generalization to all others. The paper is organized as follows.~In section \ref{intro} the basic notations and conventions of Clifford algebras are introduced used throughout this paper.~In section \ref{clif} the  real vector space of hyperconics  is introduced and identified with a real vector space by means of the mapping $\tau$. This  allows us to identify the space of hyperconic sections ${\Co}_2$ with  the set of symmetric matrices. The Clifford algebra  ${\Cl}({\Co}_2)$  is then defined. The decision hypersphere  is briefly recalled in section \ref{ellip} as a concept naturally introduced in conformal space  which is used in  defining the spherical perceptron. This leads us to define the concept of {\it elliptical perceptron} used throughout the paper as a special case of the spherical perceptron.~In  section \ref{decispr} we state  as lemma \ref{Lema1} the embedding  $  \imath :\Rr^m \hookrightarrow M^s$  and we introduce the embedding  $\Rr^m \hookrightarrow {\Cl}({\Co}_2)$. This allows us to characterize in lemma \ref{bas}  the elementary but basic incidence property of a point lying on a hyperconic  using only the Clifford product.  We state and recall briefly the one to one correspondance between the space of conics in $\Pn_2$, the set of hyperplanes in $\Pn_5$ and the dual projective space $\Pn_5^{\ast}$.
The  definition of the $d$-uple embedding $\rho_d$  is briefly recalled and used  for the special case of $d=2$ to conclude that
 the relation of a point $x \in \Pn_2$ being incident to a  plane conic is equivalent to find a hyperplane  $ \Pn_5$ containing  $\rho_2(x)$.~We solve the problem of determining the boundary decision  hyperplane by means of duality in Clifford algebra in proposition \ref{vec}.
 Duality in projective geometry and in ${\Cl}({\Co}_2)$  are equivalent in the sense of remark~\ref{gen} using corollary~\ref{uni} to proposition~\ref{vec}.~We relate the mappings $ \tau, \rho_2$ and $ \imath $ previously introduced by means of proposition \ref{theo}. The relationship between  proposition \ref{theo} and the definitions given by  ~\cite{PF04} is stated in remark \ref{tau}.~In section~\ref{exp}  the experimental results are given by  producing input data  for $m=2$ training the elliptical perceptron
as a  neural network by means of the backpropagation algorithm. The results are stated in table~\ref{conics}.~The final conclusions are stated in section~\ref{conc}.


\sect{Clifford Algebras and the Clifford Algebra for  the vector space of conics.}\label{intro}
In this section we recall the basic notation, facts  and well known properties of Clifford algebras, for a more comprehensive treatment we refer the reader
 to~e.g.,chapter 15 of \cite{Po95} or chapters 3 and 4 of \cite{PL97}  and we will restrict ourselves to introduce   the main notions and notational conventions used throughout this paper.~We denote an $m$-euclidean  vector space as ${\Rr}^m$ with its usual quadratic form. In ${\Rr}^m$ we fix as basis $e_1, e_2, \ldots, e_m $ and  denote by ${\Cl}( \Rr^m)$ or simply  ${\Cl}_m$~~if it is clear that we are forming the Clifford algebra over the real field,  the Clifford algebra associated to the $m$-dimensional euclidean quadratic space. A hypersurface of degree 2 in ${\Rr}^m$ will be called a hyperconic section or hyperconic.~If the Clifford algebra is to be emphasized   associated to the quadratic space we enclose it within  parenthesis as for example ${\Cl}(\Rr^m)$. The Clifford algebra  ${\Cl}_m$~as a real vector space has dimension $2^m$. Considering  the usual embedding of ${\Rr}^m$ in ${\Cl}_m$ and denoting by the same symbols the vectors $e_1, e_2, \ldots, e_m$ under this embedding these are called the {\it basis blades}. For mathematical applications , it is equally valid   and useful to introduce the geometric algebra ${\Ge}_{p,q,r}$ as the geometric algebra of dimension  $2^m$ where $ m = p+q+ r$ which is defined from its  underlying vector space ${\Rr}^{p,q,r}$ endowed with a signature $(p,q,r)$ by application of a geometric product.~In the sequel,  we will only consider non-degenerate geometric algebras ${\Ge}_{p,q}$ where $r=0 $. Besides, we will write ${\Ge}_m$ if $q = 0 $. In particular, note that $ {\Cl}_m = {\Ge}_m $ with this notation. Another example
is projective space which is ${\Ge}_{3,1}$.~Points in this space are represented by  $1$-blades.
 The geometric product of two multivectors $\bld a$ and $\bld b$ is simply denoted by   ${\bld a \bld b}$.~The geometric product consists of an outer  product ($\wedge$) and an inner  product ($\cdot $ ).~More precisely, as ${\Ge}_m$ is generated as an ${\Rr}$-algebra by its basis blades,  the geometric product of two basis vectors is given by :
$$
e_i e_j \stackrel{\rm def}{=} \left \{
                    \begin{array}{cc}
                     1   & \mbox{for}~~~  i=j \in \{1, \ldots, p \}, \\
                   -1  & \mbox{for}~~~  i=j  \in \{ p+1, \ldots, p+q\}, \\
                     0   & \mbox{for}~~~ i=j  \in \{ p+q+1, \ldots,  m\}, \\
  \,\,\,\,e_{ij} = e_i \wedge e_j = -e_{ji} & \mbox{for}~~i \neq j .
                     \end{array}
            \right.
$$
                                                    
 The outer product is a special operation  defined within Clifford algebra and is equivalent to the exterior product of  the Grassmann algebra. It is associative and distibutive. For vectors $ {\bld x}, {\bld y }\in {\Rr}^m $ it is also anti-commutative, i.e~ $ {\bld x} \wedge {\bld y}  = - {\bld y} \wedge {\bld x} $.~Another important property is that for a set $ \{ {\bld x}_1, \ldots, {\bld x}_k \} \subset {\Rr}^m $ of $ k \leq m$ mutually linearly independent vectors, $ {\bld x}_1 \wedge {\bld x}_2 \cdots \wedge {\bld x}_k  \wedge {\bld y}  = 0 $ if and only if $ {\bld y} $ is linearly dependent with respect to $\{ {\bld x}_1, \ldots, {\bld x}_k \}$. The   outer product of $k $ vectors is called a $k$-{\it blade} and is denoted by
$$
A_{<k>} = {\bld a}_1 \wedge {\bld a}_2 \cdots \wedge {\bld a}_k \stackrel{\rm def}{=} \bigwedge_{i=1}^k {\bld a}_i
$$
The {\it grade} of a blade is simply the number of  vectors that ``wedged'' together  give the blade. Hence, the outer product of $k$ linearly independent vectors gives a blade of grade $k$,~i.e.~ a $k$-blade. The {\it unit pseudoscalar} of ${\Cl}_m$ is a blade of grade $m$ with magnitude $1$ and denoted by $I$.~In   geometric algebra, blades, as defined above, are given a geometric interpretation. As for example the 1-blades  are the vectors, the 2-blades or bivectors are the oriented planes and so on. This is also based on their interpretation as linear subspaces. For example, given a vector ${\bld a }\in {\Rr}^m$, we can define a function ${\Ok}_{\bf a}$ as 

\begin{eqnarray*}
{\Ok}_{\bf a}:& {\Rr}^m \rightarrow  &   {\Cl}_m \\
        {\bld x} & \mapsto    & {\bld x} \wedge {\bld a}
\end{eqnarray*}        
       
\noindent The kernel of this function  is called the outer product null space (OPNS)  of 
${\bld a} $  and denoted by $\No ({\bld a} ) $.~We can explicitely describe it as:
$$
\No = \{ {\bld x} \in {\Rr}^m: {\bld x} \wedge  A_{<k>} = 0 \}.
$$
 Therefore  the OPNS of the vector ${\bld a} $ is a line through the origin with the direction given by ${\bld a }$.  In general, 
the OPNS of some $k$-blade $ A_{<k>} \in {\Cl}_m $ is a $k$-dimensional linear subspace of ${\Rr}^m $.
Another  useful concept we will use is  the null space of blades with respect to the inner product denoted as  the inner product null space (IPNS) of a blade $ A_{<k>}$, denoted by 
${\Ni}( A_{<k>})$ which is  defined as the kernel of the function

\begin{eqnarray*}
 {\In}_{A_{<k>}}:{\Rr}^{m}  & \rightarrow  & {\Cl}_m  \\
  {\bld x} & \mapsto     &  {\bld x} \cdot A_{<k>}. 
\end{eqnarray*}    
     
\noindent   which is given explicitely as ${\Ni}(A_{<k>}) = \{ {\bld x} \in  {\Rr}^m : {\In}_{A_{<k>}}( {\bld x}) = 0 \}$.
An important  notion is the dual operation  in the Clifford algebra. The dual of a multivector $ A \in {\Cl} $, denoted as $ A^{\star}$ is defined as $ A \cdot I^{-1} = A I^{-1}$ where   $I^{-1}$ is the 
inverse unit pseudoscalar, which is also an $m$-blade.~A property  useful relating both $\Ni$ , $\No$ which will be used  in  proposition~\ref{vec} is the following:
\begin{Lemm}
\label{dua}
 For a $k$-blade $ A_{<k>}$ :
 $$
  {\No} ( A_{<k>} ) = {\Ni} (A_{<k>}^{\star}). 
  $$
  \end{Lemm} 
  {\it Proof}~:~ According to ~equation (3.34) of \cite{Co00} if  $C,\, B_{<l>} $ are a $1$-blade (resp. an $l$-blade): $ (C \wedge B_{<l>})^* = C \cdot ( B_{<l>})^* $  for $ l \le m-1$ which 
  gives directly the $ ``\, \subset \, "$ contention. As for the other set theoretical contention, the last equation gives in fact $ (C\wedge A_{<k>})^*= 0 $ for    $ C \in
  {\Ni} (A_{<k>}^{\star})$ hence $ C \wedge A_{<k>}I^{-1} =  ( C\wedge A_{<k>})I^{-1} = 0$ multiplying by $I$ gives in the ${\Rr}$-algebra: $1(C\wedge A_{<k>}) = 0 $. \Fin   
  \begin{Ex}
 The OPNS of a bivector in ${\Rr}^3$ is  the IPNS of the  cross product of its vectors, a nice property only valid for three euclidean vector space.
 \end{Ex}
  The projective space $ {\Ps}^m $ is  the $m+1$ dimensional vector space ${\Rr}^{m+1}$ without the origin. In conformal geometric algebra ${\Ge}_{4,1}$ the spheres are the basis entities from which the other entities are involved , see ~e.g.\S 3 of \cite{PF04}.~Even though we will work in the sequel with  the conformal space of 3-dimensional Euclidean space, all formulae extend directly to $m$-dimensions.  In order to obtain a conformal space, the euclidean  $m$-space 
  ${\Rr}^m$ is embedded in conformal space denoted by ${\mathbb K}^m$ via  the stereographic projection  and  this space will be denoted by ${\Pk}^m$. To obtain a basis, we extend the orthonormal basis $\{ e_1,  \ldots, e_m \}$ of ${\Rr}^m$ by two orthogonal basis vectors 
 $\{ e_+, e_{-}\}$ with $ e_{+}^2 = - e_{-}^2 = 1$.
 
 A set of geometric entities of  interest in computer vision are conic sections. It is therefore useful to construct the Clifford algebra over a  real vector space such that the conics and their incidence properties such as the union, intersection etc; can be  represented in terms of the INPS and the ONPS as represented above. The idea of using the Clifford algebra for the vector space of conics has already been introduced by ~e.g. \S 4 of \cite{PF04}. The authors use this idea  to express the classical problem of fitting a set of given 
 points in ${\Rr}^2$ to a real conic and also to fit a set of conics as given input data to a 
 cluster of points in a least square sense ( see \cite{ChL02} for a recent survey of the methods to investigate this problem).
 
 \subsection{The Clifford Algebra for the real vector space of hyperconic sections} \label{clif} 
 
 It is well known from linear algebra that  for a symmetric $ 3 \times 3 $ matrix $A$  the set of  vectors  $ x= (x_1,x_2,1) $ that satisty $ x^t A x = 0 $ where ${}^t$ denotes the transpose of a vector, lie on a conic containing the point $(x_1, x_2)$. One then says that  $A$ represents   the conic defined by the equation above.~More generally,  we introduce the following vector spaces  precisely as:
 $M \stackrel{ \rm def} {=} M_{m,m}(\Rr) $ , the  space of real $m$ by $m$ matrices and $ {M^s} \stackrel{\rm def}{=} \{ A \in M | \, A = A^t \}$ the subvector space of {\it symmetric}  $m $ by $m$ matrices of $M$. We can identify  the first with the 
 $  m^2$ dimensional vector space $ {\Rr}^{m^2}$ by means of the isomorphism
 $\tau: M  \rightarrow   {\Rr}^{m^2} $ given explicitely by:
 \begin{eqnarray*}
  ( x _{i,j})_{ \, i,j \in \{ 1, \ldots, m \}} & \mapsto &      ( x_{1,m}, x_{2,m},\ldots, x_{m-1,m},{x_{m,m} \over \twoi},                                  {x_{1,1} \over \twoi},{ x_{2,2} \over \twoi}, x_{1,2}, {x_{3,3} \over \twoi}, x_{2,3}, x_{1,3} ,\\
 &   &       ,\ldots,{x_{m-1,m-1} \over \twoi}, x_{m-2,m-1}, \ldots, x_{1, m-1}). 
\end{eqnarray*}
   Note  that to describe such an isomorphism we are choosing a special permutation of the orthonormal basis of ${\Rr}^{m^2}$  followed by   a homothety.~One reason for choosing such a special permutation is because of remark~\ref{tau} in subsection~\ref{decispr}.
In particular  for  an element $A \in M^s$ :
\begin{eqnarray*}
\tau (A) &   =   &    ( a_{1,m}, a_{2,m}, \ldots, a_{m-1,m},   { a_{m,m} \over \twoi} ,  {a_{1,1}\over \twoi}, {a_{2,2}\over \twoi}, a_{1,2},{a_{3,3}
\over \twoi},  a_{2,3}, a_{1,3}, {a_{4,4}\over \twoi}, \\
 &  &    a_{3,4}, a_{2,4}, a_{1,4}, \ldots,  {a_{m-1,m-1}\over \twoi}, a_{m-2,m-1}, \ldots, a_{1,m-1} ) 
 \end{eqnarray*}
 \noindent which implies  that $ {\tau|}_{M^s} = \tau {\scriptstyle \circ} \jmath $    where $ \jmath :M^s \hookrightarrow M $ is the  inclusion. In the sequel, we will adapt the shorter notation $ \tau|$ for the restriction  instead of writing the full formula. We will use  the proof of the following
\begin{Lemm} \label{mats}
  There is an isomorphism
  $ \tau|: M^s \simeq {\Rr}^{ \bhalf m(m+1)} $ of~~$\Rr$-vector spaces.
\end{Lemm}
{\it Proof}~:~The dimension of both vector spaces are equal, so it is enough to show injectivity or surjectivity. We show the latter. If $ r= (r_1, \ldots, r_N ) \in {\Rr}^{N}$ where $ N ={\bhalf m(m+1)}$ we will define the following matrix:
$$
 R = \left ( \begin{array}{cccccccc}
{\twoi } \, r_{m+1}  &  r_{m+3}  & r_{m+6} & r_{m+10} & r_{m+15} & \cdots & r_{N-1} & r_1 \\
 r_{m+3}       & {\twoi} \, r_{m+2}  & r_{m+5}  & r_{m+9} & r_{m+14} & \cdots  & r_{N-2} & r_2     \\
 r_{m+6}& r_{m+5}  &   {\twoi} \, r_{m+4}  & r_{m+8} & r_{m+13} & \cdots & r_{N-3} & r_3  \\
 r_{m+10} &  r_{m+9} & r_{m+8}   & {\twoi} \, r_{m+7}   &r_{m+12}    & \cdots &   r_{N-4}    & r_4  \\
\vdots & \vdots &  \vdots      & \vdots  & \vdots & \vdots & \vdots & \vdots \\
\vdots & \vdots & \vdots  & \vdots & \vdots & \vdots  & \vdots  & \vdots \\
r_{N-1} & r_{N-2}     & r_{N-3}   & r_{N-4}    &\cdots     & \cdots    &      {\twoi} \, r_{{ m^2 -m +4 \over 2} } & r_{m-1} \\
 r_1 &  r_2     &  r_3  &  r_4   & \cdots    & \cdots    &  r_{m-1} & {\twoi} \, r_m  
  \end{array}
  \right).
  $$
  It is  clear that $\tau|(R) = r $ as required.
  \Fin

 As a consequence of Lemma ~\ref{mats}  the space of {\it hyperconic sections} denoted as ${\Co}_2 $  is represented by the $ {\half m(m+1)} $-dimensional euclidean vector space   and its Clifford algebra  is denoted by  ${\Cl}( {\Co}_{2})$; we will study some  of its properties of incidence in subsection ~\ref{decispr}.
  In the sequel  if $ x \in {\Rr}^{m} $ we will
denote by $ x' = (x,1) \in {\Rr}^n $, where $m = n-1 $. It will be useful  to  introduce the following: 
\begin{Def}  \label{embed}
 Let 
 $ {\imath} : {\Rr}^{m} \rightarrow    M^s   $ be defined as 
 $x    \mapsto      {x'}^t\, x'$ 
 where  the product in the right is the usual matrix product.
 \end{Def}

 \sect{ A Decision boundary hyperplane  in ${\Cl}( {\Co}_2)$ for  ${\Rr}^{\bhalf m(m+1)}$.}
 \subsection{The decision hypersphere for ${\Pk}$, the spherical perceptron and the elliptical perceptron.} \label{ellip}
\begin{figure}[htbp]
\begin{center}
\includegraphics[width=6cm]{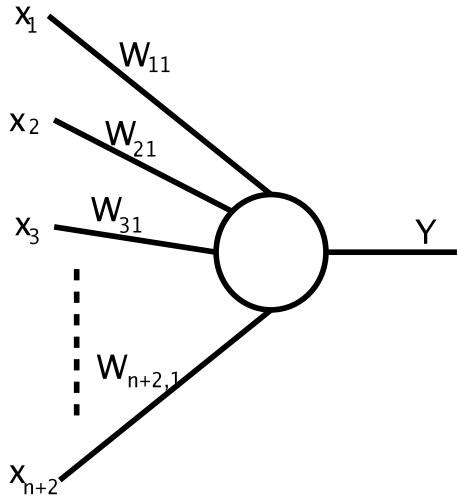}
\end{center}
\caption{\label{spercep}{\textit{Spherical  Perceptron.}}}
\end{figure}

 It is well known that Clifford algebra   is used to represent geometric entities like lines and planes through the origin in ${\Cl}_3 $.  Conformal space extends this idea by embedding  the $m$-dimensional Euclidean space  as a regular map ( in the projective-geometric sense) in an $ m+2$-dimensional space. Conformal space derives its name from the fact that certain types of reflections in conformal space represent inversion in Euclidean space and conformal transformations can be represented as  compositions of inversions in the sense of affine geometry. We have already introduced  the conformal space ${\Pk} $. The embedding of a euclidean vector ${\bld x}$ in conformal space is  given by 
$$
{\bld X} =  {\bld x} + \half{\bld x}^2 e_{\infty} + e_{o}  
$$
where $ e_{\infty}\stackrel{\rm def}{=} e_{-}+ e_{+}$ and $e_{o}\stackrel{\rm def}{=} {\half( e_{-} -e_{+})}$. Using the null basis $\{ e_{\infty}, e_{o} \} $ instead of $\{ e_{+}, e_{-} \}$ leads to the representation  of  $e_{\infty}$(resp.~$e_{o}$) as the point at infinity (resp.~the origin).~A vector of the form 
$ {\bld S}= {\bld X}-{ \half} {\rho}^2 e_{\infty} $ represents a sphere centered on ${\bld x}$ with radius ${\rho} $ and in higher dimensions represents a hypersphere. A decision hypersphere has the property that it separates points of the input data into points outside and inside the sphere; such a decision hypersphere has been determined  and is given as:
$$
\frac{{\bld S} \cdot {\bld X}}{({\bld S}\cdot e_{\infty}) ({\bld X}\cdot e_{\infty})}
 \left \{  \begin{array}{ccc}
         >  & 0   & :{\bld x} \mbox{ inside sphere},\\
         =  & 0   &: {\bld x} \mbox{ on sphere}, \\
         < & 0    &: {\bld x} \mbox{ outside sphere}
           \end{array}
        \right.     
 $$ 
whenever  $  {\bld X} \in {\mathbb H}^{3}_{a} $ where $ {\mathbb H}^{3}_{a} $ is the  affine null cone (see e.g.~equation 2.26 of \cite{PHi04}).~The significance of the affine null cone  is that it represents the vectors in $\Pk_m$ whose $e_o$ component is unity. The decision hypersphere allows us to define the  {\it spherical perceptron}  represented in figure~\ref{spercep} which shows that  it has $m+2$   weights  $w_{ij}$ and  $m+2$ inputs $x_i$ and one  output function $y $. As a special case  we define the {\it elliptical perceptron} as the spherical
perceptron with   6 weights, 6 inputs  and one output function.  
\subsection{Boundary Decision hyperplanes using  duality in Projective Geometry and in ${\Cl}( {\Co}_{2})$.}
\label{decispr}
~Given a set of input data, to classify the set of two classes of points a decision hyperplane is determined  whenever  a linear one is possible. In order to determine   the boundary  decision hyperplane we are posing  the problem of determining the boundary of the decision hyperplane. In this section we will determine one,   using the concept of duality in projective geometry and relate it to its dual in ${\Cl}( {\Co}_{2})$. In the sequel, recall that $m = n-1$.

\begin{Lemm}
$\imath $ embeds ${\Rr}^{m}$ into $M^s$. 
\label{Lema1}
\end{Lemm}
{\it Proof}~:~ Explicitely 
$$
\imath(x) =  \left( \begin{array}{ccccc}
                   x_1^{2}  & x_1 x_2 & \cdots & x_1 x_{m} & x_1  \\
                   x_2 x_1   & x_2^2   & \cdots & x_2 x_{m} & x_2 \\
                     \vdots  & \vdots &  \vdots  &\vdots &  \vdots \\
                x_{m}x_1 & x_{m}x_2 & \cdots & x_{m}^2 & x_{m} \\
                x_1  &  x_2  & \cdots  &  x_{m}   & 1 
                \end{array}
               \right)
$$
and  observe that the $m+1$-th column completely determines $ \imath $.
\Fin 

\begin{Ex}. Note that  $\imath$  can be extended to ${\Rr}^{m+1}$  but it is no longer an embedding. For example $ \imath (x_1, x_2, x_3 ) = \imath ( -x_1, -x_2, -x_3) $.
\end{Ex}
A direct calculation shows that for $ x \in {\Rr}^{m}$:
\begin{eqnarray*}
\tau (\imath(x)) &   =   &    ( x_{1}, x_{2}, \ldots, x_{m},   \sqtwo ,     \sqtwo x_{1}^2, \sqtwo x_{2}^2, x_{1} x_{2}, \sqtwo x_{3}^2,  x_{2}x_{3}, x_{1}x_{3}, \sqtwo x_{4}^2, \\
 &  &    x_{3}x_{4}, x_{2}x_{4}, x_{1}x_{4}, \ldots,   \sqtwo x_{m}^2, x_{m-1}x_{m}, \ldots, x_{1}x_{m} ). 
 \end{eqnarray*}
 
 \noindent In particular for $ m=2$ the above formula reduces for $ x = ( x_1, x_2, 1)$ to: 
 $$  
 \tau(\imath(x)) = ( x_1, x_2, \sqtwo, \sqtwo x_{1}^{2}, \sqtwo x_{2}^{2}, x_{1} x_{2} ). 
 $$
 \begin{Def} \label{hyper}
  Let  $ \underline{x} = \tau \imath(x) $ ,  $x \mapsto \underline{x} $ defines an embedding of ${\Rr}^{m}$ into ${\Co}_2 \hookrightarrow {\Cl}({\Co}_2) $.
  \end{Def} 
 \begin{Lemm}
 \label{bas}
 Denote by $\cdot$ the dot  product in ${\Cl}( {\Co}_2) $ and for  $ A \in M^s$ 
 let $ a = \tau(A) $ then:
 \begin{eqnarray*}
 \lefteqn{\underline{x} \cdot a = 0} \\
  &  \Longleftrightarrow & x_1^2 a_{11} + x_2^2 a_{22} +\cdots  + x_{m}^2 a_{m, m} + \\
  &   & + \,2 x_1 x_2 a_{12} + 2x_1x_3 a_{13} + \cdots + 2 x_1 x_{m} a_1 a_{m} +  \cdots +  a_{m+1,m+1} = 0 \\
  & \Longleftrightarrow & x'^{t} A x' = 0 .
  \end{eqnarray*}
\end{Lemm}
{\it Proof}~:~This is a direct calculation and follows from the definitions.\Fin
 
 \noindent  Note that by lemma~\ref{bas} in order to test if a conic defined by $a = \tau(A)$ contains a point $ x$ it is enough to  test whether  its clifford product $ \underline{x} \cdot  a $ is zero or not.
 \begin{Ex} 
For ~ $ m=2$ , $\underline{x} = ( x_1, x_2, 1) $:
 \begin{equation}
 \label{dot}
\underline{x}\cdot a = 0   \Longleftrightarrow  x_1^2 a_{11} + x_2^2 a_{22} + 2 x_1x_2 a_{12} + 2x_1a_{13} + 2x_2 a_{23}  + a_{33} = 0.
\end{equation}
where again $\cdot $ is the  dot product in ${\Cl}( {\Co}_2)$.
\end{Ex}
The set of conics in the two dimensional projective space ${\Pn}_2$  in the homogeneous coordinates $ (x:y:z) $ is  given by:
$$
{\cal C} = \{ x^2 a_{11} + y^2 a_{22} + z^2 a_{33} + 2xy a_{12} + 
 2 xz a_{13} + 2 yz a_{23} = 0 \}
 $$ 
 If we introduce coordinates $( a_{11}: a_{22}: a_{33}: 2 a_{12}: 2 a_{13} : 2 a_{23} ) $    for $ {\Pn_5}^{\ast} $  then we can define a one-to-one correspondence:
 ${\cal C } \longleftrightarrow {{\Pn}_5}^{\ast}$ given as $  c \mapsto ( a_{11}: a_{22}: a_{33}: 2 a_{12}: 2 a_{13} : 2 a_{23} ) $. 
 
 It is well known , see ~e.g. exercise 2.12, chapter I of \cite{Ha97},~  from the projective geometric properties of regular maps of${\Pn}_2$ that there is  a regular mapping, which is an embedding,   the so-called  {\it d-uple embedding} $\rho_d$  which considers all  monomials of degree $d$ in the variables $x_0, \ldots, x_{m} $,which are $ {m+d \choose  m}$ and substituting  each  homogeneous coordinate of the  point $P= (a_0: \ldots :a_m) $  in the monomials thus giving a map $ \rho_d: \Pn_m \rightarrow \Pn_M $ where $ M = {m+d \choose m} -1 $. 
\begin{Ex} 
If   $m=1$, $ d=2 $  the double embedding of $\Pn_1$ in $\Pn_2$ has as image a conic curve.
\end{Ex}
~Another typical  example  is given by $ m= 2$, $ d=2$ the image $\rho_2( \Pn_2)$ is a surface  called the {\it  Veronese surface}.~Let $N = {\half (m+1)(m+2)} $ and  for the application of $\rho_d$ to the case of hyperconics,~$ d=2$ and  $ M = { m+2 \choose 2} -1 = N-1$.~
  
As a consequence of  Lemma~\ref{bas}  is that  for   $ x = ( x_1: x_2: 1)$ , $ \rho_2 (x) \in \Pn_5 $  and a hyperplane $H_{x}$ in $\Pn_5 $  containing  this point is given by Eq.~ (\ref{dot}). 
 Duality in projective geometry  is an  isomorphism of projective spaces which defines for each hyperplane $H$ in ${\Pn}_5 $ as above, a point $ ( a_{11}: a_{22}: 2 a_{12}: 2 a_{13}:
 2 a_{23}:  a_{33}) \in {\Pn}_5^{\ast}$  and conversely, for each  $ a \in {\Pn_5}^{\ast} $ corresponds a unique hyperplane in ${\Pn}_5$, namely the hyperplane  $ H_a$ defined by the equation:
 $$
  z_1a_{11} + z_2a_{22} +  z_3 a_{12} +  z_4a_{13} +  z_5 a_{23} + z_6a_{33}  = 0 .
  $$
 ~Note that the point $a$ defined in lemma \ref{bas} is up to an automorphism the point  in the veronese surface.~We conclude  from the previously stated  isomorphisms $ {\cal C} \longleftrightarrow {\Pn_5}^{\ast} \longleftrightarrow      \{ \mbox{hyperplanes in} \,\, \Pn_5 \} $ that to find a conic in 
 $\Pn_2$ containing $x$ it is    sufficient to  find a point in $ {\Pn_5}^{\ast}$ or equivalently a hyperplane $H_x$  in ${\Pn}_5 $  containing $\rho_2(x)$.
 
 We solve the  problem   of determining the boundary decision hyperplane   by means of duality in Clifford algebra. More precisely,
\begin{Prop}
\label{vec}
Let $ x^{(1)}, \ldots ,x^{(N)}$ define a set of  mutually  linearly independent vectors in  ${\Rr}^m $ and let  $ u = \und{x}_1 \wedge \ldots \wedge  \und{x}_N $ where $\und{x}_i = \tau( \imath(x^{(i)} ))$ for $ i = 1, \ldots, N $  then 
$\No ( u ) = \Ni (u^{\ast}) $ where $ u^{\ast}$ is unique up to a constant.
\end{Prop}
{\it Proof}:~ Each vector $ \und{x}_i$ is linearly dependent with $ u $ hence ~$u \wedge \und{x}_i = 0$ for  all $i$. Hence by lemma \ref{dua},
$u^{\ast} \cdot \und{x}_i  = 0 $ for all $i$. Hence $ u^{\ast} \in W$, where 
$ W = < \und{x}_1, \und{x}_2, \ldots, \und{x}_N >^{\perp}$ which is one-dimensional.   
\Fin
\begin{Cor} \label{uni}
With the same hypothesis as lemma~\ref{vec},  $u^{\ast}$ is the unique conic which incides  through the points $x^{(1)}, \ldots, x^{(N)}$.
\end{Cor}
{\it Proof}~:~This follows inmediately from lemma \ref{vec}. 
\Fin

Duality in projective geometry and duality in  Clifford algebra  are equivalent in the following sense: 
\begin{Obs}
\label{gen}
Let  $m, N$ as before and $ x \in \Rr^{m} $.~To find a hyperplane $H$ in  $ {\Pn}_N $ containing 
${\rho_2}(x)$ it is sufficient to find $N $ points: $ x^{(1)}, \ldots, x^{(N)}$ mutually linearly independent in $\Rr^{m}$ which determine the $N$-blade 
$ x = \und{x}_1\wedge  \ldots \wedge \und{x}_N $ and its Clifford dual $ x^{\ast}$ which is a vector in $ {\Cl}({\Co}_2)$. In homogeneous coordinates it is the  projective dual to the hyperplane $H$.
\end{Obs}

The simplest case of remark~\ref{gen} is the following: 
\begin{Ex}
For $m= 2, N = 5 $ and $ x \in {\Rr}^2 $,  to determine a hyperplane $H$ in ${\Pn}_5 $ containing ${\rho}_2(x)$  it is enough to find five points no three of which are collinear. Denoting by $ x^{(1)}, \ldots, x^{(5)}$ these points and by  
$ x = \und{x}_1 \wedge \ldots \wedge \und{x}_5$ , $x^{\ast}$ is a vector in 
${\Cl}( {\Co}_2 )$  which is an element of $ {\Pn_5}^{\ast}$.
\end{Ex}
Let $ {\Af}_{M}  \stackrel{\rm def}{=} \{ x \in {\Pn}_M | x_{m+1} = 1 \} $ ,  ${\Af}_m \stackrel{\rm def} {=} \{ x \in {\Pn}_m | x_{m+1} = 1 \} $ and 
\begin{eqnarray*}
 {\rho}_2 |:{\Af}_n  &  \rightarrow   &  {\Af}_{M} \\
   x          &    \mapsto  & 
  [ x_1: x_2: \ldots: x_{m}: 1: x_1^2 : x_2^2: x_1 x_2 : \ldots: x_1 x_{m} ]. \\
 \end{eqnarray*}
 which is the restriction of the double embedding.
 Let 
  $$
    s(l) = \left \{  \begin{array}{cc}
                     m+1 &   l = 0 ,    \\
                      m+2  &  l = 1 ,    \\
                      s(l-1) + l-1,  &  2 \le l \le m .
                      \end{array}
      \right.                
  $$
  Note that in particular $ s(m) = {\scriptstyle \frac{(m+1)^2 -(m+1) +4}{2}}$. The integers $ \{ s(i) \}_{i = 0}^{m} $ define a set with $m+1$ elements $S$.
 Define the following mappings:   
  \bas
    T: {\Af}_M \rightarrow {\Af}_M & , & \{ x_i \}_{ i=1}^N \mapsto \left \{  \begin{array}{cc}
                         \sqrt{2}  x_i    & i  \in S  \\
                          x_i           & \mbox{otherwise.}
                          \end{array}
      \right. ,   \\
  p: \Rr^N -\{ 0 \}  \rightarrow  {\Af}_M &, &  ( x_1, \ldots, x_N) \mapsto (x_1: \ldots : x_N) ,\\  
   q: {\Af}_m \rightarrow {\Rr}^m & , &(z_1: \ldots: z_{m}:1) \mapsto (z_1, \ldots,z_m).  
  \eas 
  The relation  between all the maps above is given by the following:
\begin{Prop}
\label{theo}
The following diagramme:
$$
\def\normalbaselines{\baselineskip20pt \lineskip3pt \lineskiplimit3pt }
\def\mapright#1{\smash{\mathop{\longrightarrow}\limits^{#1}}}
\def\mapleft#1{\smash{\mathop{\hbox to 100 pt{\leftarrowfill}}\limits^{#1}}}
\def\mapdown#1{\Big\downarrow \rlap{$\vcenter{\hbox{$\scriptstyle#1$}}$}}
\begin{array}{cc}
{\Pn}_m \supseteq {\Af}_m & > \hspace{-9pt}>\hspace{-7pt}\mapright{q} {\Rr}^m \stackrel{ \imath}{\hookrightarrow} M^s \stackrel{\tau |}{\hookrightarrow}{\Rr}^N -\{ 0 \} \\
    \mapdown{ \rho_2 |}    &   \hspace{40pt}\mapdown{p}  \\
    \Pn_M \supseteq {\Af}_M&  \mapleft{T}\hspace{-6pt}<\hspace{-9pt}< {\Af}_M  \subseteq {\Pn}_M
    \end{array}
 $$ 
 is conmutative, where the open ended arrows are  isomorphisms, $ \rho_2 |$  is only  an embedding  and  $p$  is only surjective.~More precisely,   $  T {\sci} \, p \, {\sci}\, (\tau|) \, {\sci} \,{\imath}\, {\sci}\, q = \rho_2 | $.
 \end{Prop}
 {\it Proof}:~~This is a direct consequence of the definitions of the mappings given above.  
 \Fin

 \begin{Ex}
  For the space of  plane conic sections $ d=m= 2$, $M= 5 $.~By fixing an ordering on the monomials,    $ \rho_2 : {\Pn}_2 \hookrightarrow {\Pn}_5 $  is the mapping $ (x_1:x_2: x_3) \mapsto ( x_1^2: x_2^2:x_3^2: x_1 x_2: x_1 x_3: x_2 x_3) $.~For this case, the double embedding  is defined at the corresponding affine charts $ {\Af}_2 = \{  x \in \Pn_2 |  x_3 = 1 \} $ , $ {\Af}_5 = \{ x \in {\Pn}_5 |x_3 = 1 \} $ given by its restriction:
 $ \rho_2 | : {\Af}_2 \rightarrow {\Af}_5 $; in this case  $ S =\{ 3,4,5 \} $ and $T: {\Af}_5 \rightarrow {\Af}_5 $ is the  automorphism  given by:
 $$
 (\xi_1: \xi_2: \xi_3: \xi_4: \xi_5: \xi_6) \mapsto ( \xi_1: \xi_2: \sqr \xi_3: \sqr \xi_4: \sqr \xi_5: \xi_6 ).
 $$
 Note that the inverse image of a hyperplane of $\Pn_5$ under $\rho_2$ is a conic in $\Pn_2$.
  \end{Ex} 
\begin{Obss}\label{tau}
The authors in ~\cite{PF04} introduce ${\Cl}({\Co}_2)$ only for the case of plane conic sections and define the mappings  ${\cal T}$ and  ${\cal D}$ stating  no  apparent relation amongst these mappings. In our case $ {\cal T} = \tau$ and ${\cal D}(x) = \underline{x}$ in our notation hence stating their close relationship. ~We complete the relation amongst these mappings   by introducing the mappings $p, T $ and $\rho_2|$ which is   summarized by  prop.~\ref{theo}.
\end{Obss}

\subsection{Experimental results to determine the boundary decision hyperplane.} 
\label{exp}

\begin{figure}[htbp] 
\begin{center} (a)\includegraphics[width=6cm]{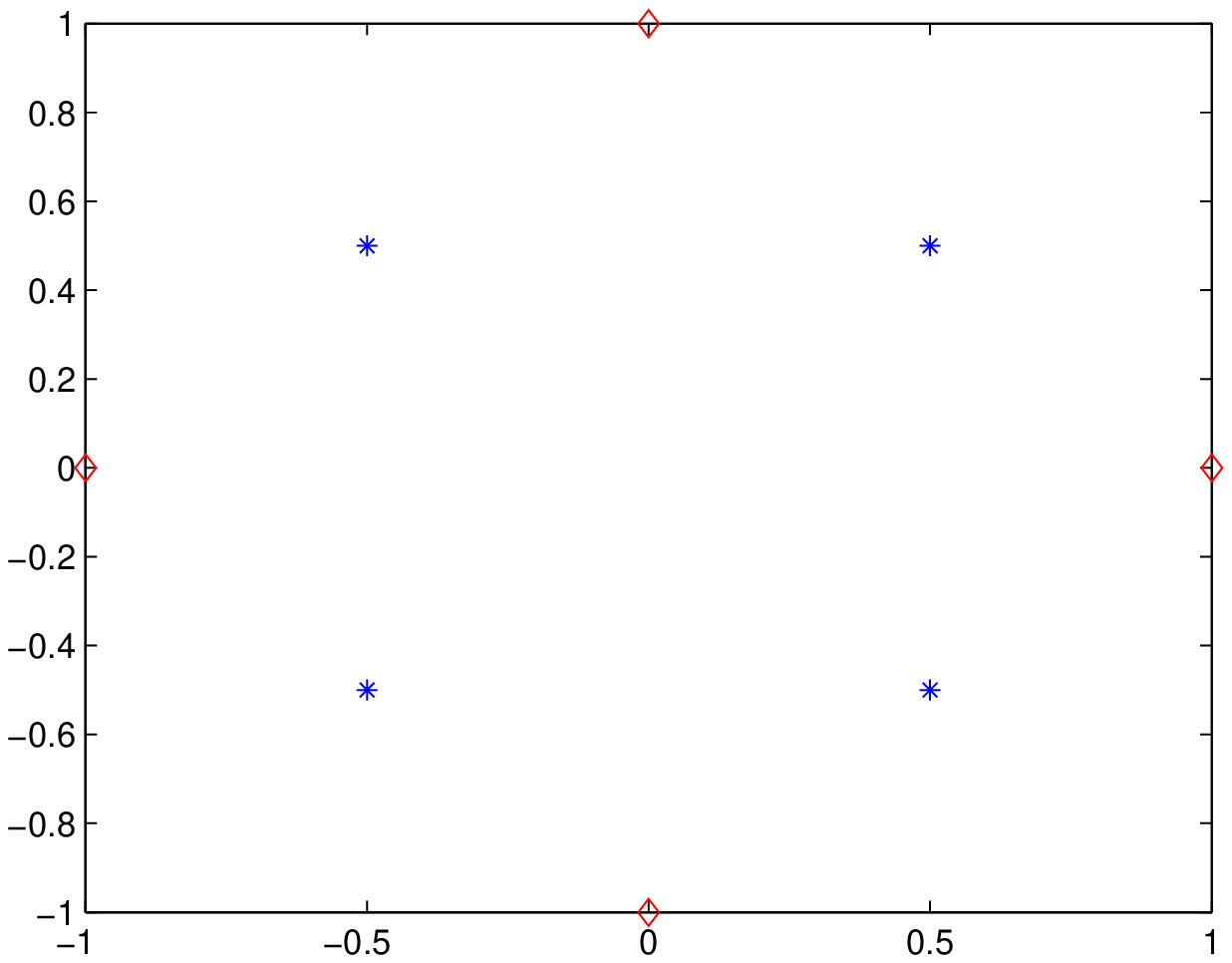} (b)\includegraphics[width=6cm]{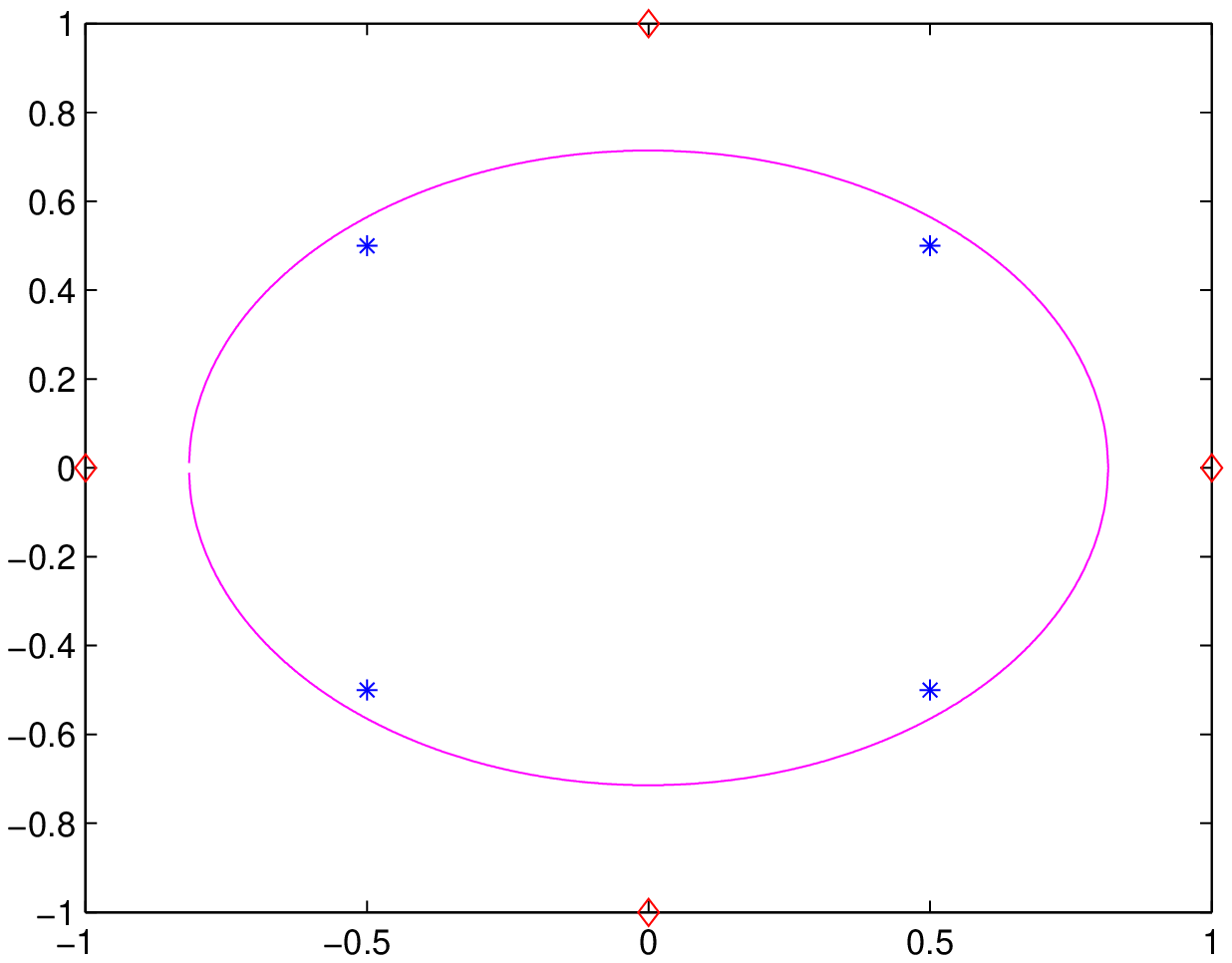}
              (c)\includegraphics[width=6cm]{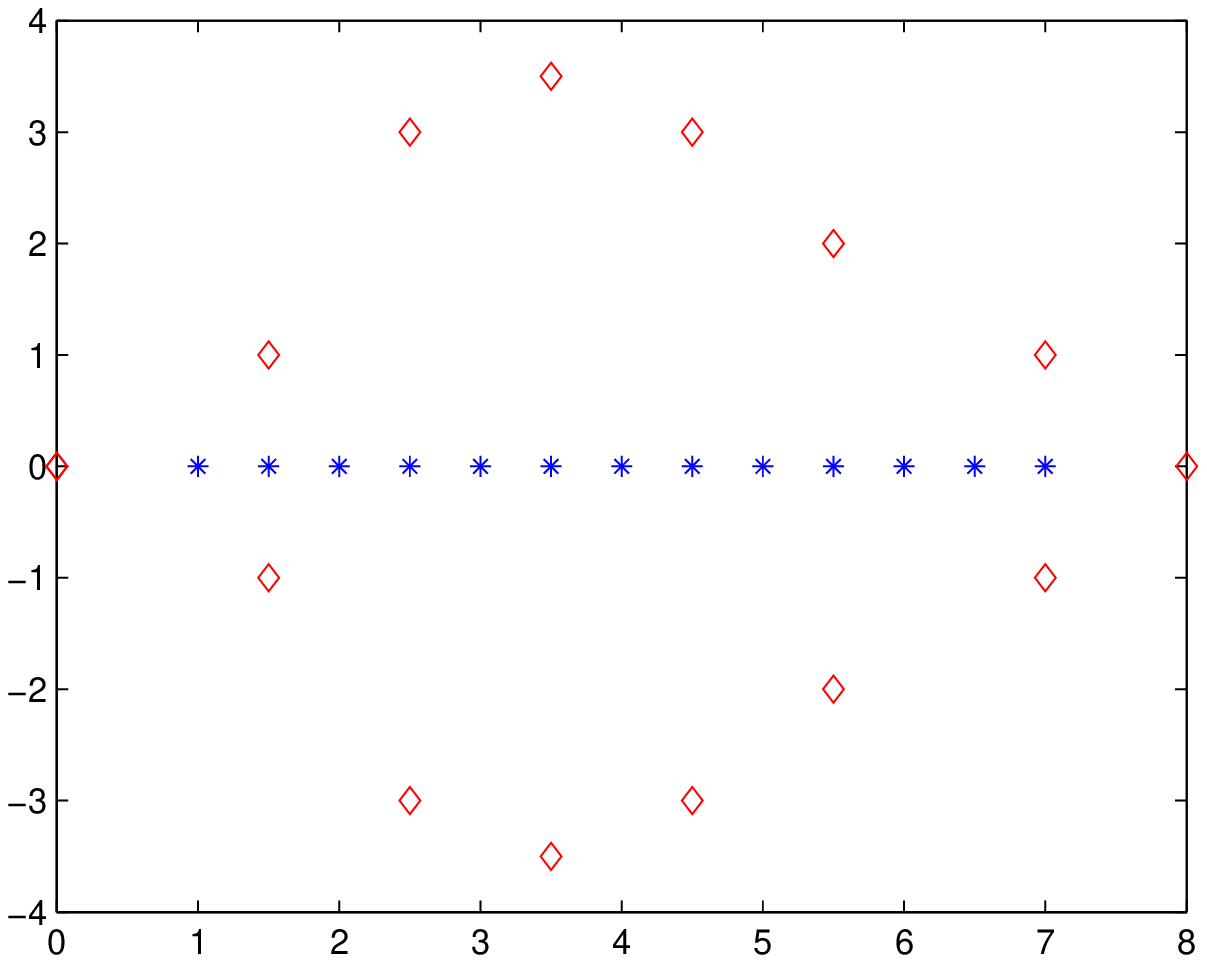} (d)\includegraphics[width=6cm]{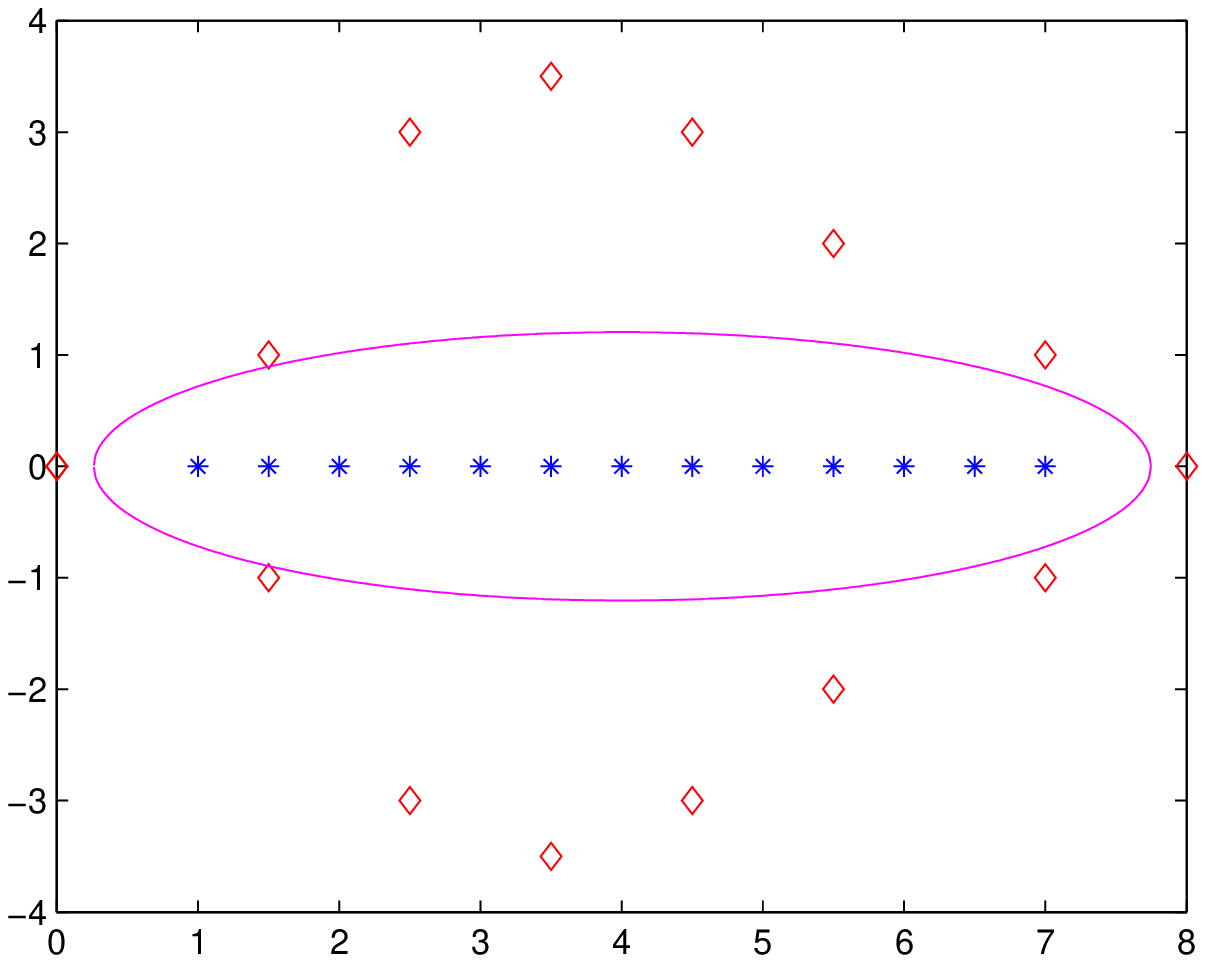}

   (e)\includegraphics[width=6cm]{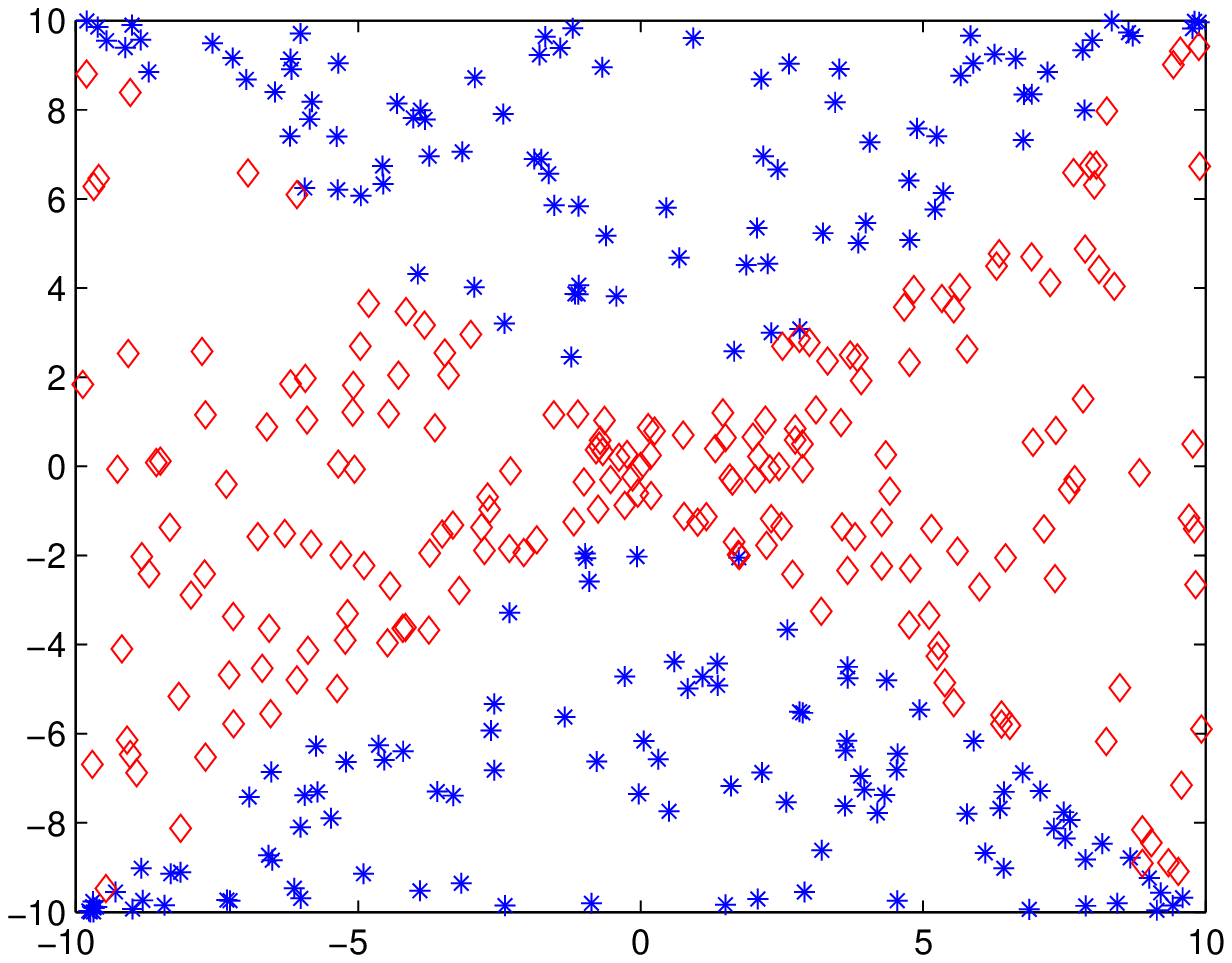} (f)\includegraphics[width=6cm]{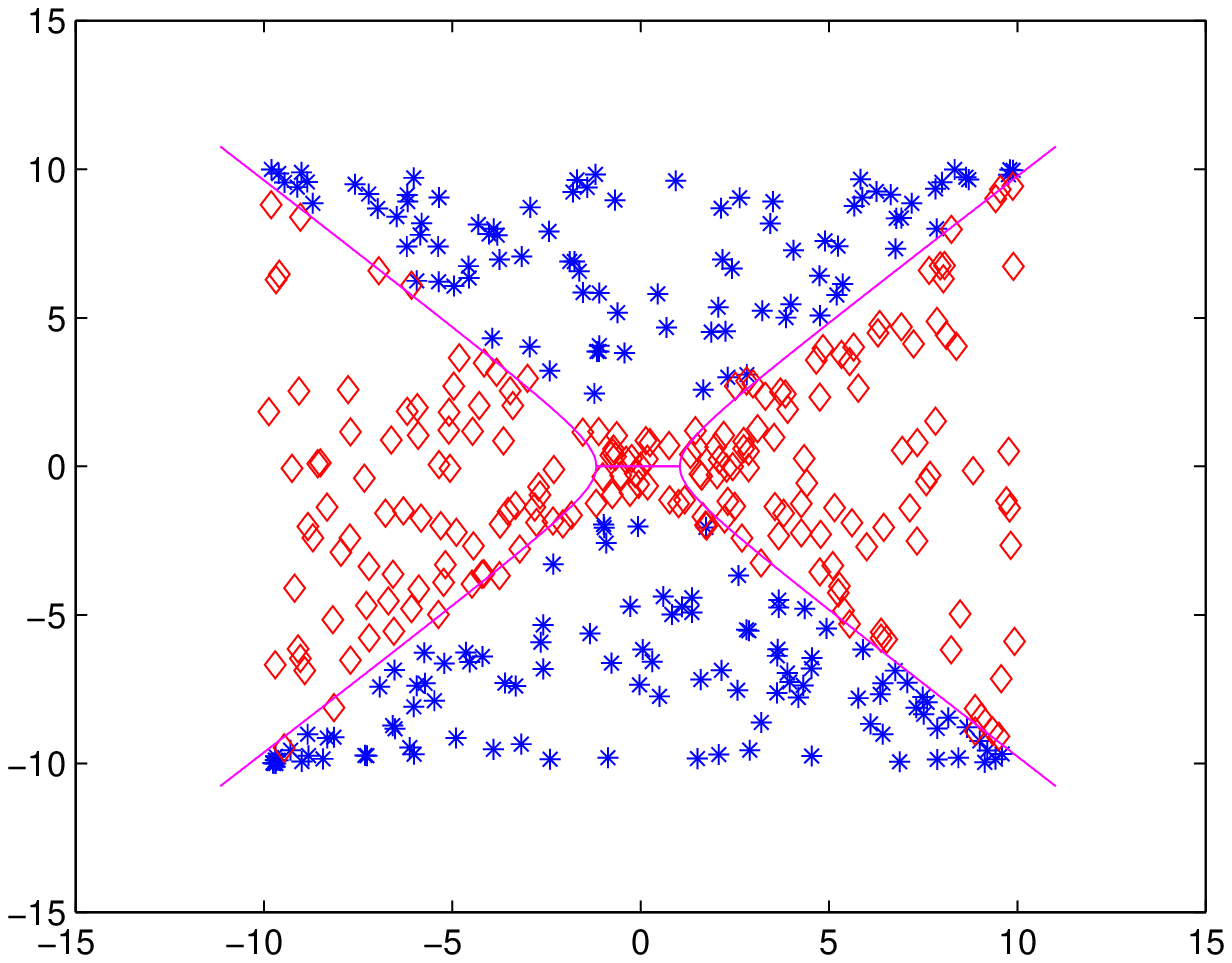}
\end{center}
\caption{\label{dibuj}{\textit{Points to separate (a), (c) and (e)  and decision conic (b),(d) and (f)}}}
\end{figure}

 In order to obtain experimental  results we produced data of points  for $m=2$, that is to say  plane conics in $\Rr^2$.~The decision hyperconic  is  to be determined by using the {\it elliptical perceptron} defined at the end of subsection \ref{ellip}  which   has weights $ \{ \omega_{i} \}_{i=1}^6$ and  with $6$ inputs and one  output function. 
Each of the examples considered for the elliptical perceptron is tabulated in table~\ref{conics} given below. We give in each case as data for the MLP  a set of points divided in two classes to be separated by a decision boundary hyperplane. For the data,  to train the neural network  6 nodes  for the input and one node for the output with no hidden layers in both cases were chosen. The learning rule is  the backpropagation algorithm where the input function was chosen to be the dot product with typical   transfer functions as the sigmoid bipolar and the sine bipolar to properly  bound the output in the interval $[-1,+1]$.~In order to obtain the equation of the conic we obtained  the set of  weights $\omega_1, \ldots, \omega_6$. If we let $ \omega = ( \omega_1, \ldots, \omega_6)$ using  $\tau$ :
 $$
  \tau^{-1}( \omega) =  \left( \begin{array}{ccc}
  \sqr \omega_4  &   \omega_6   &  \omega_1      \\
   \omega_6     & \sqr \omega_5  & \omega_2     \\
  \omega_1    &  \omega_2        & \sqr \omega_3                    
              \end{array}
               \right)
$$ 
and the equation of the conic in this case is:
$$
 \sqr x^2 \omega_4 +  \sqr y^2 \omega_5 + 2 xy \omega_6 + 2 x \omega_1  + 2 y \omega_2  +  \sqr \omega_3 = 0.
 $$
This equation was then tranformed into the standard form to obtain the equation of the estimated conic described in the last column of table~(\ref{conics}) for each vector  $\omega$ of weights.
\begin{table}[htbp]
   \begin{tabular}{|l|c|r|} \hline
 Conic &  Weights  $(\omega_{1},\ldots,\omega_{6})$ & Estimated Conic Equation \\
     \hline \hline
     Ellipse & $(0.00,0.00,-3.30,5.00,6.36,0.00)$ & $\frac{x^2}{0.66}+\frac{y^2}{0.51}=1$  \\
     \hline
     Ellipse &$(8.48,0.00,-2.84,-1.50,-14.43,0.00)$&$\frac{(x-4.005)^2}{14.075}+\frac{y^2}{1.45}=1$  \\
     \hline
   Hyperbola & $(-2.23,0.00,-8.26,-19.05,20.2,0.00 )$ &$\frac{(x+0.07)^2}{1.23}-\frac{y^2}{1.17}=1$  \\
     \hline
\end{tabular}
\center
 \caption{\label{conics} Results for the experimental points.}
\end{table} 
  In  figure~\ref{dibuj} we graph in the first column the two classes of points to be separated for each of the examples of table~\ref{conics}. The first class of points is denoted by a cross and the second by a diamond. A decision boundary hyperplane is to be determined in $\Rr^2$. In the second column the decision conic is drawn, showing  the separation between both classes of points.

\sect{Conclusion}\label{conc} 
The elliptical perceptron introduced in this paper generalizes the spherical perceptron  used in conformal geometry to determine the boundary  decision hypersurface in euclidean $m$-dimensional space. We have shown that, by means of Clifford algebra the usual space of hyperconic sections embeds into the Clifford algebra of hyperconic sections; this allows us to  use all the properties of the geometric product enjoyed by this Clifford algebra and as we have shown, also  the Clifford Dual is essential to determine the vector orthogonal to the boundary of the decision hyperplane.~A projective property of the space of hyperconics  is that it is equivalent to the set of hyperplanes in the projective dual and then it is  proved  that for each such hyperplane its  orthogonal vector is in fact  the Clifford dual since to find a decision boundary hyperplane in  the  euclidean $m$-dimensional space, it is enough in terms of the space ${\Cl}({\Co}_2)  $ to  determine an  $m-1$-blade generated by  $m-1$ pairwise independent vectors and  evaluate its Clifford dual which  is {\it a fortiori} the orthogonal vector to the original hyperplane.~In the experiments to  test the theory introduced in subsection~\ref{clif} to determine a boundary decision hyperconic we linearize the problem of finding the  hyperconic section by embedding the input  data by means of the double-embedding $\rho_2$.~The MLP of the elliptical perceptron is introduced to determine  a  vector orthogonal to the hyperplane in this feature space and then the inverse mapping $ \tau^{-1}$  is applied  to the vector. We then use  equation \ref{dot} for this special case to evaluate the equation of the estimated conic.~Note that the procedure we have outlined is completely general and does not depend  on the dimension of the ambient input space.~The experimental results in subsection \ref{exp} are only done for typical examples  which is for  plane conics, where it is shown that  there exists one decision boundary  conic for each of the input data  given in table~\ref{conics}.~By training the elliptical perceptron the estimated  vector orthogonal to the boundary of the decision hyperplane is evaluated.~Using  $\tau^{-1} $~ the estimated equation of the conic is computed.~This procedure might at first hand seem very special but  the theory developed so far can be done  is developed  for  the  higher dimensional case  as  the maps $\rho_2$ and $\tau$ are completely independent of  the dimension of the ambient  space  and the typical examples in such cases will then be the more general   hyperconics sections,   where  again  a vector orthogonal to the boundary decision hyperplane needs to be determined by  exactly the same procedure  and $\tau^{-1}$ is   used to determine the equation of the estimated general hyperconic and only the values for $n,m, N$ have to be once again determined.

\bibliographystyle{unsrt}

\end{document}